\documentclass{article}
\usepackage{arxiv}

\usepackage[utf8]{inputenc} 
\usepackage[T1]{fontenc}    
\usepackage{hyperref}       
\usepackage{url}            
\usepackage{booktabs}       
\usepackage{amsfonts}       
\usepackage{nicefrac}       
\usepackage{microtype}      
\usepackage{lipsum}		
\usepackage{graphicx}
\usepackage[sort, numbers]{natbib}
\usepackage{doi}
\usepackage{multirow}
\usepackage{multicol}
\usepackage{array}
\usepackage{amsmath}
\usepackage[ruled,vlined]{algorithm2e}
\usepackage{subfigure}

\title{A Dual Cross-Attention Graph Learning Framework For Multimodal MRI-Based Major Depressive Disorder Detection}

\author{ \href{https://orcid.org/0000-0000-0000-0000}{\includegraphics[scale=0.06]{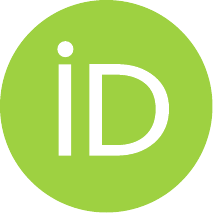}\hspace{1mm}Nojod M. Alotaibi*}\\
        Department of Computer Science, \\
        Faculty of Computing and Information Technology,\\
        King Abdulaziz University, Jeddah, Saudi Arabia\\
        \texttt{nalotaibi0351@stu.kau.edu.sa}\\
	\And
	\href{https://orcid.org/0000-0000-0000-0000}{\includegraphics[scale=0.06]{orcid.pdf}\hspace{1mm}Areej M. Alhothali*}\\
        Department of Computer Science, \\
        Faculty of Computing and Information Technology,\\
        King Abdulaziz University, Jeddah, Saudi Arabia\\
        \texttt{aalhothali@kau.edu.sa}\\
}

\renewcommand{\shorttitle}{\textit{arXiv} Template}

\hypersetup{
pdftitle={A template for the arxiv style},
pdfsubject={q-bio.NC, q-bio.QM},
pdfauthor={David S.~Hippocampus, Elias D.~Striatum},
pdfkeywords={First keyword, Second keyword, More},
}

\begin{document}
\maketitle

\begin{abstract}
	Major depressive disorder (MDD) is a prevalent mental disorder associated with complex neurobiological changes that cannot be fully captured using a single imaging modality. The use of multimodal magnetic resonance imaging (MRI) provides a more comprehensive understanding of brain changes by combining structural and functional data. Despite this, the effective integration of these modalities remains challenging. In this study, we propose a dual cross-attention–based multimodal fusion framework that explicitly models bidirectional interactions between structural MRI (sMRI) and resting-state functional MRI (rs-fMRI) representations. The proposed approach is tested on the large-scale REST-meta-MDD dataset using both structural and functional brain atlas configurations. Numerous experiments conducted under a 10-fold stratified cross-validation demonstrated that the proposed fusion algorithm achieves robust and competitive performance across all atlas types. The proposed method consistently outperforms conventional feature-level concatenation for functional atlases, while maintaining comparable performance for structural atlases. The most effective dual cross-attention multimodal model obtained 84.71\% accuracy, 86.42\% sensitivity, 82.89\% specificity, 84.34\% precision, and 85.37\% F1-score. These findings emphasize the importance of explicitly modeling cross-modal interactions for multimodal neuroimaging-based MDD classification.
\end{abstract}


\section{Introduction}
\label{sec:introduction}
Major Depressive Disorder (MDD) is one of the most prevalent and debilitating mental disorders in the world, affecting millions of individuals and causing significant social and economic burdens~\cite{cui2024major}. Although there is extensive clinical research, the current diagnosis is based primarily on symptom-based assessments and subjective clinical interviews, which can lead to inconsistencies in diagnosis and delayed interventions. However, several neuroimaging studies have shown that it is closely related to structural and functional abnormalities in specific brain regions~\cite{pilmeyer2022functional}. In the last decade, neuroimaging has emerged as a promising avenue to identify potential biomarkers of MDD~\cite{tulay2019multimodal}. Modalities such as structural magnetic resonance imaging (sMRI) and resting-state functional MRI (rs-fMRI) provide complementary insights crucial to the development of automated diagnosis frameworks for a wide range of brain disorders~\cite{tulay2019multimodal}. sMRI provides high-resolution structural information of the brain, reflecting anatomical and morphological alterations related to gray matter (GM) structure~\cite{zhuo2019rise}. sMRI studies revealed structural changes in many brain regions in patients with MDD, including the temporal lobe, frontal lobe, and limbic system~\cite{dai2019brain}. In contrast, rs-fMRI measures boldoxygen-level-dependent (BOLD) signals of subjects to detect abnormalities or dysfunction on the functional connectivity network (FCN)~\cite{pilmeyer2022functional}. The rs-fMRI findings found that MDD patients had abnormal FC related to the central executive network (CEN), the salience network (SN), and the default mode network (DMN)~\cite{dai2019brain}.

Currently, most neuroimaging-based MDD studies have focused on examining structural or functional imaging biomarkers through machine learning (ML) and deep learning (DL) approaches that can analyze complex medical images and assist physicians in making accurate diagnoses. Several sMRI-based studies, including~\cite{ar20253d},~\cite{gao2023classification},~\cite{alotaibi20263dvit}, and~\cite{xiao2024diagnosis}, have demonstrated that structural changes provide discriminative information for MDD. Similarly, rs-fMRI-based studies such as~\cite{alotaibi2025multi},~\cite{xia2023depressiongraph},~\cite{lee2024spectral}, and~\cite{liu2023spatial} have reported significant functional abnormalities associated with the disorder. Despite their respective strengths, unimodal investigations do not take advantage of the complementary information provided by combining sMRI with rs-fMRI. 

To overcome this limitation, recent studies have increasingly explored multimodal learning frameworks that jointly analyze sMRI and rs-fMRI data. Such multimodal neuroimaging approaches have emerged as one of the most promising research fields in neuroscience~\cite{zhang2020advances}. It has been demonstrated in several multimodal studies that combining sMRI and rs-fMRI can improve MDD classification performance compared with unimodal models~\cite{mousavian2021depression, zheng2023attention, yuan2023cross, fan2025classifying}. In this context, graph-based learning methods have attracted significant interest, as they model connectivity patterns among brain regions in neuroimaging data. Graph Neural Networks (GNNs) offer a powerful framework for representing relationships between brain regions, where nodes correspond to regions of interest (ROIs) and edges describe structural or functional interactions~\cite{wang2022adaptive}. In particular, GNN-based methods effectively capture inter-regional dependencies and exhibit strong performance in neuroimaging-based disease classification~\cite{alotaibi20263dvit}. Despite these advances, many multimodal MDD studies combine sMRI and rs-fMRI features using simple concatenation strategies, which fail to capture the relationships between structural and functional brain networks. In some graph-based multimodal studies, structural and functional brain graphs are constructed and processed separately. Consequently, cross-modal interactions between these modalities remain largely unexplored. To the best of our knowledge, bidirectional attention mechanisms for modeling interactions between sMRI and rs-fMRI graphs have not been examined in MDD studies.

Furthermore, effective feature representation plays a crucial role for improving neuroimaging-based classification performance. Traditional ML and DL approaches often utilize local feature extraction mechanisms that may fail to capture complex spatial dependencies and long-range relationships among brain regions. Recently, Vision Transformers (ViTs) have demonstrated strong capability in medical image analysis by capturing global contextual relationships through self-attention mechanisms~\cite{dai2024classification}. Several studies have reported that ViT-based models achieve improved performance compared with conventional convolutional neural networks (CNNs) in MRI-based disease classification tasks~\cite{bi2023multivit, shaffi2024ensemble}.

To address these shortcomings, we propose a dual cross-attention–based multimodal fusion framework for MDD detection that integrates sMRI and rs-fMRI data to exploit complementary brain information. The main contributions of this work are summarized as follows:
\begin{enumerate}
    \item A ViT-based structural representation module that extracts high-level ROI embeddings from 3D sMRI data and constructs a structural brain graph to model anatomical relationships between brain regions.
    \item A functional graph construction strategy that models inter-regional interactions using functional connectivity estimated from rs-fMRI signals.
    \item A unified Graph Attention Network (GAT) encoder that learns node embeddings from both structural and functional graphs within a consistent architecture.
    \item A dual cross-attention fusion mechanism that captures bidirectional interactions between structural and functional representations, enabling each modality to refine the features of its counterpart before final classification.
\end{enumerate}

\section{Related Work}
\label{sec:RelatedWork}
In recent years, advances in neuroimaging-based MDD analysis have increased research efforts on multimodal learning approaches that leverage complementary information from sMRI and rs-fMRI data. These methods can capture richer representations of disease-related brain changes compared to single-modality models. With the advent of deep learning and graph-based representations, multimodal frameworks are evolving from simple feature fusion approaches to more expressive architectures that explicitly model cross-modal interactions. In this review, we summarize representative studies in the field and place the proposed method within a broader framework of multimodal neuroimaging studies.

Early multimodal studies relied primarily on handcrafted connectivity or radiomic features coupled with classical machine learning techniques. For instance, Mousavian et al.~\cite{mousavian2021depression} developed an ML workflow based on brain connectivity features extracted from sMRI and rs-fMRI images using spatial cube similarity measures. Their method assessed multiple ML classifiers and produced an accuracy of 91\% on the NKI dataset. Similarly, Ma et al.~\cite{ma2022gray} employed radiomic features extracted from sMRI and diffusion tensor imaging (DTI) with a Random Forest classifier (RF), reporting 86.75\% accuracy in distinguishing MDD from healthy controls (HCs). However, these methods are highly dependent on manual feature engineering and feature selection procedures, which may limit their scalability and ability to capture complex multimodal feature interactions.

Graph-based DL methods have also been extensively investigated. Pan et al.~\cite{pan2022mamf} designed a multi-scale adaptive multi-channel fusion GCN (MAMF-GCN) framework using attention mechanisms for predicting mental disorders. Several channels were used to build phenotyping data and fMRI's population graphs under different brain atlases. Their approach achieved an accuracy of 97.67\% on the Autism Brain Imaging Data Exchange (ABIDE) dataset and 99.25\% on the MDD dataset. Wang et al.~\cite{wang2022adaptive} developed the Adaptive Multimodal Neuroimage Integration (AMNI) framework combining Graph Convolutional Network (GCN) for rs-fMRI and 3D Convolutional Neural Network (CNN) for sMRI, with an adaptation module to reduce modality discrepancy. The AMNI algorithm was evaluated on the REST-meta-MDD dataset and achieved an accuracy of 65\%. Although graph-based multimodal frameworks enhance connectivity modeling, structural and functional graphs are typically created separately and fused using feature aggregation or attention mechanisms, which may not adequately capture reciprocal cross-modal interactions at the embedding level.

Recently, several studies have focused on the use of attention mechanisms and transformer-based architectures to improve the integration of multimodal information. Zheng et al.~\cite{zheng2023attention} proposed an attention-based multimodal framework integrating structural and functional features, achieving 72.00\% accuracy on the REST-meta-MDD dataset. Yuan et al.~\cite{yuan2023cross} developed the Brain Dynamic Attention Network (BDANet), which generates dynamic brain graphs and analyzes them using a GCN. Moreover, the model integrated an ensemble classifier to mitigate multisite domain variability, yielding an overall accuracy of 81.6\%. Similarly, Chen et al.~\cite{chen2025mmdd} introduced the Multimodal Multitask Dynamic Disentanglement (MMDD) framework, in which sMRI features were extracted using a 3D ResNet backbone and functional time-series data was modeled using an LSTM-Transformer encoder. Further, a bidirectional cross-attention fusion mechanism was integrated into the framework, providing 77.76\% accuracy on the REST-meta-MDD dataset. 
 
Meanwhile, domain-adaptive and federated multimodal approaches have been introduced to address multisite heterogeneity. Fan et al.~\cite{fan2025classifying} developed the personalized Federated Gradient Matching and Contrastive Optimization (pF-GMCO) algorithm to deal with domain shift and facilitate scalable MDD classification using multimodal MRI data. Their model utilized gradient matching using cosine similarity to weight contributions obtained from different sites, contrastive learning for client-specific model optimization, and multimodal compact bilinear (MCB) pooling to combine sMRI and fMRI features. The pF-GMCO model achieved 79.07\% accuracy on the REST-meta-MDD dataset. Nevertheless, this framework predominantly focused on federated adaptation and client-specific optimization, with limited emphasis on explicitly modeling structured cross-modal interactions at the representation level.

To overcome these limitations, we propose a dual cross-attention mechanism that operates directly on structural and functional brain graphs. Specifically, bidirectional cross-attention is applied between corresponding nodes of the two modality-specific graphs, enabling reciprocal refinement of node embeddings while maintaining graph topology. By explicitly modeling structural–functional dependencies at the node level, the proposed approach enables more effective integration of complementary information and improves discriminative representation learning for robust MDD classification. 

\section{Dataset Description}
\subsection{REST-meta-MDD Dataset}
As part of this study, we used data from the REST-meta-MDD consortium, a publicly available and open-access repository that represents the largest multi-site dataset for major depressive disorder (MDD) to date~\cite{chen2022direct}. This dataset consists of 2,428 subjects collected from 25 imaging sites, including 1,300 MDD patients (826 females and 474 males) and 1,128 HCs~\cite{chen2022direct, yan2019reduced}. 

In each site, comprehensive phenotypic information was collected, including age, sex, episode status (first-episode or recurrent), medication status, illness duration, and symptom severity assessed using the 17-item Hamilton Depression Rating Scale (HAMD)~\cite{chen2022direct}. Furthermore, the dataset includes two imaging modalities for each participant: T1-weighted structural MRI (sMRI) and resting-state functional MRI (rs-fMRI). Data collection procedures at each contributing site were approved by the local Institutional Review Boards, and all participants provided written informed consent before participating~\cite{chen2022direct, yan2019reduced}. 
\subsection{Data Preprocessing}
The structural MRI (sMRI) and resting-state functional MRI (rs-fMRI) data were preprocessed using standardized neuroimaging pipelines to maintain data quality and consistency between subjects. T1-weighted sMRI images were preprocessed using the Computational Anatomy Toolbox (CAT12) implemented in SPM12~\cite{yan2019reduced}. The preprocessing pipeline involved bias-field correction, skull stripping, and segmentation into gray matter (GM), white matter (WM), and cerebrospinal fluid (CSF)~\cite{yan2019reduced}. The segmented images were then spatially normalized to the Montreal Neurological Institute (MNI) standard template to ensure anatomical alignment across subjects~\cite{yan2019reduced}. In this study, GM images were used for subsequent analysis as they capture relevant structural information about brain morphology. Each GM image has spatial dimensions of $121 \times 145 \times 121$ voxels. 

An analysis of rs-fMRI data was performed using the Data Processing Assistant (DPARSF) toolbox~\cite{yan2019reduced}. The preprocessing steps included discarding the initial volumes to allow signal stabilization, correcting slice-timing and head motion. The functional images were subsequently normalized to the MNI space and spatially smoothed. Linear detrending and temporal band-pass filtering were further applied to remove low-frequency drift and high-frequency noise. In addition, nuisance covariates were regressed out to reduce non-neuronal contributions, and sites with fewer than ten subjects were excluded to ensure data reliability and consistency across centers~\cite{yan2019reduced}. 

Furthermore, our analysis of regional time series extracted using selected brain atlases found that some brain regions of certain subjects contained missing signals. Therefore, these subjects were removed from the subsequent analysis to maintain data quality. In this study, $1563$ participants were retained from $16$ sites after preprocessing, of which $810$ were patients with major depressive disorder (MDD) and $753$ were healthy controls (HC). Detailed demographic information about the
subjects included in our study is provided in Table~\ref{DemoInfo}.

To remove potential confounding factors, we first regressed out the variation between subjects in age and sex using a nonlinear Gaussian process model~\cite{lei2020integrating}. Site-related variability (also called batch effects) was then corrected using the ComBat harmonization method~\cite{el2023harmonization}. ComBat applies an empirical Bayes framework to estimate and remove site-specific effects while preserving biologically meaningful variation across subjects~\cite{johnson2007adjusting, el2023harmonization}. The Combat model can be expressed as:
\begin{equation} \label{Combat}
Y_{ij} = \alpha + X_{ij} \beta + \gamma_i + \delta_i \epsilon_{ij}
\end{equation}

Where $Y_{ij}$ indicates the time series value for subject $j$ at site $i$, $\alpha$ is the overall mean, $X_{ij}$ is a design matrix for the biological covariates (i.e., sex and age), $\beta$ reflects the regression coefficients for the covariates, $\gamma_i$ and $\delta_i$ represent the site effects (additive and multiplicative effects), and $\epsilon_{ij}$ is the residual error. This procedure improves the consistency of multi-site neuroimaging data and supports reliable downstream analysis.
\begin{table}[htbp]
\caption{Demographic Information of the 1563 Study Subjects.}
\centering
\label{DemoInfo}
\renewcommand{\arraystretch}{1.5}
\begin{tabular}{|c|c|c|c|c|c|} \hline 
 \multirow{2}{*}{\textbf{Group}} & 
 \multirow{2}{*}{\textbf{Number of subjects}}& 
 \multirow{2}{*}{\textbf{Male}}& 
 \multirow{2}{*}{\textbf{Female}} &
 \multicolumn{2}{c|}{\textbf{Average $\pm$ Standard deviation (\%)} }\\ 
 \cline{5-6}
 &&&&
 \textbf{Age} & \textbf{Education}\\
 \hline 
 \begin{tabular}{c} MDD  \end{tabular}& 
 \begin{tabular}{c}810\end{tabular}&  \begin{tabular}{c}293\end{tabular}&  \begin{tabular}{c}517\end{tabular}& \begin{tabular}{c}34.39$\pm$11.55\end{tabular}& \begin{tabular}{c}11.96$\pm$3.37\end{tabular}\\ \hline
 \begin{tabular}{c}HC  \end{tabular}&  \begin{tabular}{c}753\end{tabular}&  \begin{tabular}{c}307\end{tabular}&  \begin{tabular}{c}446\end{tabular}& \begin{tabular}{c}34.61$\pm$13.17\end{tabular}& \begin{tabular}{c}13.57$\pm$3.42\end{tabular}\\ \hline
\end{tabular}
\end{table}
\section{Methodology}
\begin{figure}[htp]
\centering
\includegraphics[height=0.93\textheight, keepaspectratio]{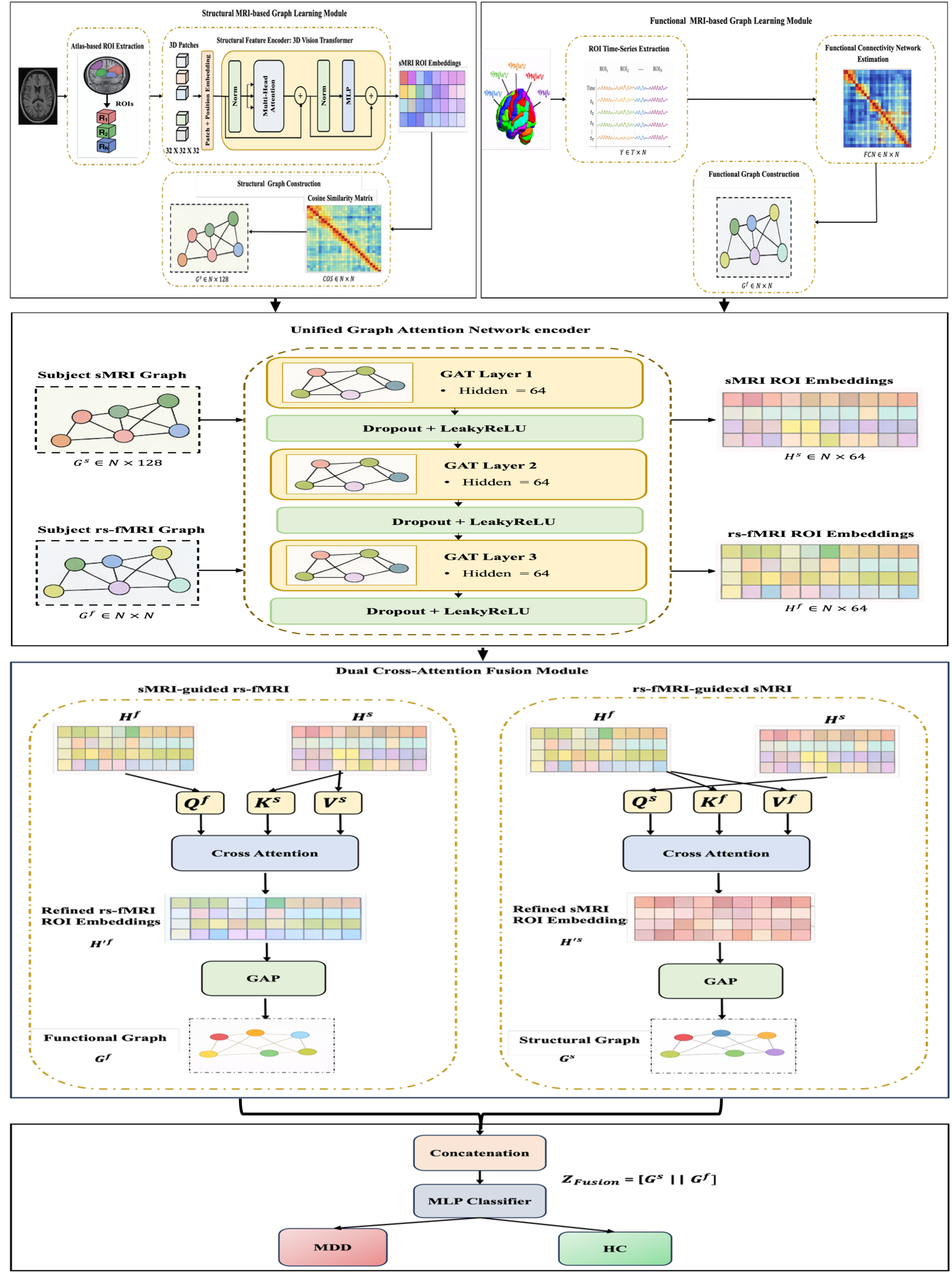}
\caption{Overall architecture of the proposed dual cross-attention multimodal framework. sMRI, structural MRI; rs-fMRI, functional MRI; ROI, brain regions of interest; MLP, multilayer
perceptron; GAT, graph attention network; LeakyReLU, leaky rectified linear unit; GAP, global average pooling; MDD, major depressive disorder; HC, healthy controls.}
\label{fig:archit}
\end{figure}

In this section, we present the proposed multimodal framework for MDD detection that integrates sMRI and rs-fMRI data from the REST-meta-MDD dataset. Our methodology extends previously developed graph-based learning frameworks for sMRI~\cite{alotaibi20263dvit} and rs-fMRI data~\cite{alotaibi2025multi}. By combining these two complementary pipelines into a single multimodal architecture, the proposed framework is designed to capture coordinated structural-functional brain alterations associated with MDD. As illustrated in Figure~\ref{fig:archit}, the overall framework is composed of four main components: (1) a structural MRI-based graph learning module for extracting anatomical features from sMRI data, (2) a functional MRI-based graph learning module for extracting functional connectivity features from rs-fMRI data, (3) a unified Graph Attention Network (GAT) encoder, and (4) a multimodal fusion module for integrating structural and functional graph representations.

\subsection{Structural MRI-based Graph Learning}
\subsubsection{Atlas-based ROI Extraction}
With the atlas-based extraction strategy, the brain is partitioned into meaningful areas of interest (ROIs) based on predefined neuroanatomical atlases. This approach is designed to represent brain organization in a structured and biologically informed manner, thereby allowing region-level analysis that is compatible with established neuroimaging studies. In this study, structural and functional atlases are used to examine the impact of different parcellation strategies. Specifically, two structural atlases: Automated Anatomical Labeling (AAL, 116 ROIs)~\cite{tzourio2002automated} and Harvard–Oxford (HO, 112 ROIs)~\cite{kennedy1998gyri} and two functional atlases: Dosenbach (Dose, 160 ROIs)~\cite{dosenbach2010prediction} and Craddock (CK, 200 ROIs)~\cite{craddock2012whole} are utilized and systematically compared. 

Atlas-based approaches define ROIs based on known anatomical or functional boundaries, as opposed to patch-based approaches that utilize uniform spatial partitioning. This produces spatially consistent regions across subjects, which reduces variability introduced by arbitrary patch placement and facilitates comparable neurobiological analysis of corresponding regions. This consistency is particularly crucial in graph-based models, since it ensures that corresponding nodes across subject-specific graphs represent the same brain regions, thereby enabling the model to learn meaningful regional patterns and retain biological interpretation.

For each subject, ROIs are extracted as volumetric patches, where the total number of patches corresponds directly to the number of regions in the selected atlas. These ROIs are represented as fixed-size 3D patches of $32 \times 32 \times 32$ voxels, ensuring uniform input dimensions across all regions and subjects. The extracted patches are then fed into the 3D Vision Transformer (ViT), which learns region-specific discriminative representations, thus enhancing model performance.

\subsubsection{Structural Feature Encoder: 3D Vision Transformer}
In recent years, the Vision Transformer (ViT) has demonstrated strong performance across a range of visual analysis tasks by modeling global dependencies through self-attention mechanisms~\cite{dosovitskiy2020image}. The ViT does not rely on local receptive fields like convolutional neural networks, but rather processes images as a sequence of patches in which long-range spatial relationships can be captured explicitly~\cite{dai2024classification}. As part of this study, we utilized a 3D ViT to extract discriminative representations from sMRI data to conduct downstream analysis.

With a preprocessed 3D sMRI volume, the image is divided into a set of non-overlapping 3D patches of size $32\times 32\times 32$. Each 3D patch represents a single ROI, which can be defined anatomically or functionally based on the chosen atlas. The resulting patch sequence is $x_{p} = [x^{1}_{p}, x^{2}_{p}, \dots, x^{N}_{p}]$, where $N$ is the number of patches (ROIs). Specifically, each 3D patch $x^{i}_{p}$ is flattened into a one-dimensional vector and then mapped to a lower-dimensional embedding space using a linear patch embedding layer~\cite{dosovitskiy2020image}:
\begin{equation}
z^{i}_0 = E x^{i}_{p}, \quad i = 1, \dots, N  
\end{equation} 

In this case, $E \in \mathbb{R}^{d \times P}$ is a learnable projection matrix, $P = 32^3$ is the patch dimensionality, and $d = 128$ is the embedding dimension, \(z^{i}_0\in \mathbb{R}^{d}\) indicates the obtained patch embedding, and $N$ is the number of patches.

Additionally, a learnable classification token $z^{\text{CLS}}_{0}$ is prepended to the patch embedding sequence, enabling the model to capture global contextual information for downstream classification as follows~\cite{dosovitskiy2020image}:
\begin{equation}
Z^{s}_0 = [z^{\text{CLS}}_0; z^{1}_0; z^{2}_0; \dots; z^{N}_0] + E_{pos}
\end{equation}

Assuming that $E_{pos}$ is the learned positional encoding matrix for the entire input sequence. This embedded sequence is analyzed by a stack of Transformer encoder layers ($L = 6$), each consisting of a multi-head self-attention (MHSA) and a multi-layer perceptron (MLP). Further, layer normalization (LN) is applied before MHSA and MLP, while residual connections are incorporated after both operations to facilitate stable training and efficient information propagation. Transformer encoder operations are formulated as follows~\cite{dosovitskiy2020image}: 
\begin{equation}
Z'^{s}_{\ell} = \text{MHSA}\left(\text{LN}\left(Z^{s}_{\ell-1}\right)\right) + Z^{s}_{\ell-1}, \quad \ell = 1,\dots,L 
\end{equation}
\begin{equation}
Z^{s}_{\ell} = \text{MLP}\left(\text{LN}\left(Z'^{s}_{\ell}\right)\right) + Z'^{s}_{\ell} \quad \ell = 1,\dots,L 
\end{equation}

During the MHSA module, the input embeddings $Z^{s}_{\ell-1}$ are linearly projected into query, key, and value matrices as follows~\cite{dosovitskiy2020image}:
\begin{equation}
    Q = Z^{s}_{\ell-1} W_Q, \quad K = Z^{s}_{\ell-1} W_K, \quad V = Z^{s}_{\ell-1} W_V,
\end{equation}
Where $Z^{s}_{\ell-1}$ denotes the input embeddings to the $\ell-th$ encoder layer, and $W_Q$, $W_K$, and $W_V \in \mathbb{R}^{d \times d_k}$ are learnable weight matrices. The head dimension $d_k = \frac{d}{h}$, where $d$ is the embedding dimension and $h$ is the number of attention heads. In our configuration, $d = 128$ and $h = 8$, resulting in $d_k = 16$.

The attention output is then computed using scaled dot-product attention~\cite{dosovitskiy2020image}: 
\begin{equation}
    \text{Attention}(Q, K, V) = \text{softmax}\left(\frac{QK^T}{\sqrt{d_k}}\right)V
\end{equation}

Where $Q$, $K$, and $V$ refer to the query, key, and value matrices, respectively. This attention operation is performed in parallel across $h$ heads, and the outputs are combined and linearly projected to produce the final attention result. In this manner, the model is capable of simultaneously capturing diverse contextual information from multiple spatial perspectives.

In this study, the entire ViT is fine-tuned using the classification objective for learning task-specific representations. We construct the final representation by concatenating the [CLS] embedding with all ROI embeddings, rather than relying solely on the [CLS] token: 
\begin{equation}
Z^{s}_{\text{final}} = [z_L^{\text{CLS}} \, \| \, z_L^1 \, \| \, z_L^2 \, \| \,\dots \, \| \, z_L^N]
\end{equation}

In this case, $\|$ denotes concatenation and $Z^{s}_{\text{final}}$ is the resulting feature vector. Our design maintains the global contextual information captured by the [CLS] token along with the region-specific structural patterns encoded in the ROI tokens, improving the discriminative power of the learned representation. The final classification output is then produced by feeding the obtained $Z_\text{final}$ into a fully connected layer (FCN) with a softmax activation.

\begin{equation}
    \hat{y} = \text{softmax}(W Z^{s}_{\text{final}} + b)
\end{equation}

Assume $W$ and $b$ are the weight and bias of the final classification layer, and $\hat{y}$ represent the predicted class probability.

\subsubsection{Structural Graph Construction}
This section explains the procedure for constructing subject-specific brain graphs from region-level embeddings derived from the 3D ViT, as shown in Algorithm~\ref{alg:S_knngraph}. 

Each subject is represented by a graph $G_s = (V, E)$, in which each node $v_i \in V$ corresponds to the brain region (ROI) defined by the selected atlas and represented by its ViT embedding $z_i$. The edges ($E$) are determined based on the pairwise cosine similarity between the region embeddings. Cosine similarity is adopted as it measures the angular consistency between high-dimensional feature vectors while remaining invariant to their magnitudes, thereby reducing the impact of inter-subject variability and feature-scale differences~\cite{you2025semantics, 10696658}. Furthermore, it introduces no additional learnable parameters and is computationally efficient, which makes it suitable for constructing similarity-based brain graphs based on deep sMRI representations~\cite{10696658, liu2025csepc}. A cosine similarity calculation between two embeddings $z_i$ and $z_j$ is performed using the following formula~\cite{bi2024gray}:

\begin{equation} \label{eq:cosine}
    \text{Cosine\_similarity}(z_i, z_j) = \frac{z_i \cdot z_j}{\|z_i\| \|z_j\|}
\end{equation}

Here, the numerator $z_i \cdot z_j$ represents the product of $z_i$ and $z_j$, while the denominator $\|z_i\| \|z_j\|$ represents the product of the Euclidean norms of these embeddings. Notably, cosine similarity scores range from -1 to 1, where 1 denotes complete similarity, 0 denotes orthogonality, and -1 represents dissimilarity.

This produces a similarity matrix $COS \in \mathbb{R}^{N \times N}$, where $N$ denotes the number of ROIs. Then, a $K$-nearest neighbor (KNN) approach with $K = 10$ is applied to build a sparse and informative graph by preserving only the top-$K$ most similar ROIs for each node. Afterward, a weighted undirected edges are created between each node and its selected neighbors, where edge weights reflect cosine similarity values. 

Overall, the proposed graph construction procedure generates a weighted, subject-specific brain graph that emphasizes the most informative inter-regional relationships while suppressing weak or noisy connections, producing a robust and discriminative structural representation for downstream GNN–based classification.

\begin{algorithm}
\caption{Structural Graph Construction Process}
\label{alg:S_knngraph}
\KwIn{Set of ViT embeddings $Z^{s} = \{z_1, z_2, \dots, z_N\}$ for subject $s$, number of neighbors $K=10$} 
\KwOut{Subject-specific graph $G^s = (V, E)$, where $V$ is the set of nodes and $E$ is the set of edges}
\vspace{0.5em}
\textbf{Structural graph construction:\\}
  \For{each subject $s = 1,\dots,S$}{
      \textbf{1. Compute the cosine similarity matrix (COS):}\\
      \For{each pair of ROIs $(i,j)$}{
        $COS_{ij}=cosine\_similarity(z_i,z_j)$
        }
      \textbf{2. Construct KNN graph $(K=10)$:}\\
         Define node set $V = \{v_1, v_2, \dots, v_N\}$, where each node corresponds to one brain ROI\;  
         \For{each node $v_i \in V$}{
          Identify the set $\mathcal{N}_K(v_i)$ of $K$ most similar nodes to a node $v_i$ based on COS\;
              \For{each node $v_h \in \mathcal{N}_K(v_i)$, $h \neq i$}{Add undirected edge $(v_i, v_h)$ with weight $w_{ij} = COS_{ij}$
              }
         }
    }
\Return {$G^s$} 
\end{algorithm}

\subsection{Functional MRI-based Graph Learning}
\subsubsection{ROI Time-Series Extraction}
Resting-state fMRI (rs-fMRI) measures intrinsic neural activity through temporal fluctuations in the blood-oxygen-level-dependent (BOLD) signal~\cite{pilmeyer2022functional}. To generate region-wise functional representations, each preprocessed rs-fMRI volume is divided into $N$ ROIs based on a predefined brain atlas. This atlas-based decomposition provides a standardized representation of brain activity among subjects that facilitates subsequent graph construction.

For each subject, the rs-fMRI signal within each ROI is averaged over voxels to produce an ROI-wise time-series matrix:
$\mathbf{Y} = [y_1, y_2, \ldots, y_N] \in \mathbb{R}^{T \times N}$

In this case, $T$ represents the number of time points, $N$ is the number of ROIs, and $y_i \in \mathbb{R}^{T}$ corresponds to the complete rs-fMRI time series of the $i$-th ROI at $T$ time points. We used $T = 140$ time points in this study. This representation preserves temporal dynamics at the regional level and reduces noise and dimensionality compared with a voxel-based representation. 
\subsubsection{Functional Connectivity Estimation}
A functional connectivity network (FCN) describes the statistical dependence between brain activity signals measured from spatially distinct brain regions~\cite{COLLIN2016313}. Specifically, a FCN can be constructed by computing the correlation between each pair of ROIs using Pearson correlation coefficients (PC) as follows~\cite{wang2022adaptive}:
\begin{equation}
\label{eq:eqPC}
\begin{aligned}
b_{ij} &= corr(y_i,y_j) \\
&= \frac{(y_i-\bar{y}_i)^{T}(y_j-\bar{y}_j)}
{\sqrt{\left(y_i-\bar{y}_i\right)^{T}\left(y_i-\bar{y}_i\right)}
 \sqrt{\left(y_j-\bar{y}_j\right)^{T}\left(y_j-\bar{y}_j\right)}}
\end{aligned}
\end{equation}

Here, $b_{ij}\in [-1,1]$ indicates the correlation between the $i$-th and $j$-th ROIs, $y_{i}$ and $y_{j}$ are time series points for ROIs $i$ and $j$, respectively, whereas $\bar{y}_{i}$ and $\bar{y}_{j}$ represent the mean values of the corresponding ROI time series.

This procedure produces a symmetric FCN $\in \mathbb{R}^{N \times N}$, where $N$ represents the number of ROIs. Fisher’s z-transformation is then applied to FCN values to improve comparability and stabilize variance across different sites~\cite{vergara2018method}.

\subsubsection{Functional Graph Construction}
In this section, the estimated FCN is transformed into a subject-specific function brain network to enable graph-based representation learning. Graph representation enables the modeling of complex inter-regional interactions that are difficult to capture with conventional vector-based formulations. In addition, it provides a more comprehensive analysis of brain connectivity, leading to more accurate predictions in brain-related studies.

Suppose that ${G}^{f} = (V, E)$ represents the functional graph for a given subject, where $V= \{v_1, v_2, \ldots, v_N\}$ are the ROI nodes, and $E$ identifies the set of edges, while $N$ is the number of ROIS. To construct an informative and sparse topology, a KNN strategy with $K =10$ is applied to the FCN. Particularly, only the $K$ ROIs with the highest functional connectivity values are kept as neighbors of each node $v_i$. Edges are then established between node $v_i$ and each selected neighbor $v_h$, producing an undirected graph. The weight assigned to each edge is given by the corresponding FCN entry: $w_{ij} = FCN_{ij}$. In addition, the node feature vector for $v_i$ is specified as the $i$-th row of the FCN, reflecting its pairwise functional relationships with all other nodes.

This functional graph $G^f$ is subsequently used as input to the unified Graph Attention Network (GAT) encoder to learn enriched node representations. The detailed procedure of functional graph construction is summarized in Algorithm~\ref{alg:fc_knn_graph}.

\begin{algorithm}
\caption{Functional Graph Construction Process}
\label{alg:fc_knn_graph}
\KwIn{ROI time-series matrix $\mathbf{Y} \in \mathbb{R}^{T \times N}$ for subject $s$, number of neighbors $K=10$}
\KwOut{Subject-specific graph $G^f = (V, E)$, where $V$ is the set of nodes and $E$ is the set of edges}
\vspace{0.5em}
\textbf{Functional graph construction:\\}
\For{each subject $s = 1,\dots,S$}{
    Let $\mathbf{Y} = [\mathbf{y}_1, \mathbf{y}_2, \dots, \mathbf{y}_N]$, where $\mathbf{y}_i \in \mathbb{R}^{T}$\;
   
    \textbf{1. Compute functional connectivity matrix (FCN):}\\
    \For{each pair of ROIs $(i,j)$}{
        $FCN_{ij} = \mathrm{corr}(\mathbf{y}_i, \mathbf{y}_j)$\;
    }
   
    \textbf{2. Apply Fisher z-transformation:}\\
    \[
    \mathbf{Z}_{ij} = \frac{1}{2}\ln\left(\frac{1+FCN_{ij}}{1-FCN_{ij}}\right)
    \]
    Set similarity matrix FCN $\leftarrow \mathbf{Z}$\;
   
    \textbf{3. Construct KNN graph ($K=10$):}\\
    Define node set $V = \{v_1, v_2, \dots, v_N\}$, where each node represents one brain ROI\;
    \For{each node $v_i \in V_i = 1,\dots,N$}{
        Identify $\mathcal{N}_K(v_i)$, the indices of the top-$K$ largest values in FCN\;
        \For{each $v_j \in \mathcal{N}_K(v_i)$, $j \neq i$}{
            Add undirected edge $(v_i,v_j)$ with weight $w_{ij} = FCN_{ij}$
        }
    }
}
\Return $G_f$
\end{algorithm}

\subsection{Unified Graph Attention Network Encoder}
Graph Attention Networks (GATs) represent a class of Graph Neural Networks (GNNs) that incorporate attention mechanisms to adaptively weight the contributions of neighboring nodes during message passing~\cite{velivckovic2017graph}. GAT aggregates neighborhood information by assigning data-driven importance coefficients to inter-node interactions, enabling an effective representation of complex and heterogeneous relationships~\cite{velivckovic2017graph}. This property is especially useful for brain graphs, as the significance of inter-regional connections may differ across subjects and imaging modalities.

In this study, a Graph Attention Network (GAT) is used as a unified encoder to learn node-level representations from subject-specific brain graphs. The same GAT architecture is applied independently for each modality, providing a consistent graph-based representation that can be applied across different modalities. The unified GAT encoder operates on graphs derived from region-wise embeddings to produce node representations, which are later used in multimodal fusions. 

For a given subject, a graph is defined as $G^{m}= (V, E)$, where $V$ represents the set of nodes that correspond to ROIs and $E$ represents the set of edges encoding inter-regional relationships of modality $m \in \{s, f\}$. In addition, each node $v_i$ is associated with a feature vector $z^{m}_i \in \mathbb{R}^{d_m}$, obtained from the upstream modality-specific encoder. In particular, $d_s = 128$ for the structural modality, while $d_f = N$ for the functional modality.

Within a GAT layer, node features are first linearly transformed using a learnable weight matrix $\mathbf{W}^{m}_{k}$ associated with the $k$-th attention head. The attention coefficient is then computed for a target node $v_i$ and each of its neighbors $v_j \in \mathcal{N}{(v_i)}$ as follows~\cite{velivckovic2017graph}:
\begin{equation}
    e_{ij} = \mathrm{LeakyReLU}\!\left( \mathbf{a}_{k}^{T} \big[ \mathbf{W}^{m}_{k} \mathbf{z}_i^{m} \parallel \mathbf{W}^{m}_{k}\mathbf{z}_j^{m} \big] \right)
\end{equation}

Where $e_{ij}^{m}$ indicates the importance of node $v_j$ attributes to node $v_i$, the Leaky Rectified Linear Unit (LeakyReLU) is a nonlinear function, $\mathbf{a}_{k}$ is a learnable attention vector, ${.}^{T}$ is a transposition, and $\parallel$ reflects the concatenation operation.

Moreover, a softmax function is applied to normalize the coefficients over all the neighbors of node $v_i$, allowing them to be compared across nodes~\cite{velivckovic2017graph}:

\begin{equation}
    \alpha_{ij}^{m} = \frac{\exp(e_{ij}^{m})} {\sum_{v_l \in \mathcal{N}(v_i)} \exp(e_{il}^{m})}
\end{equation}

Where $\alpha_{ij}^{m}$ represents the normalized attention coefficient and $exp(.)$ is the exponential function. 

These normalized attention coefficients are used to generate an updated embedding for node $v_i$ based on the aggregated information from its neighbors~\cite{velivckovic2017graph}:
\begin{equation}
    \mathbf{h}_i^{m} = \sigma\!\left( \sum_{v_j \in \mathcal{N}(v_i)} \alpha_{ij}^{m} \mathbf{W}^{m}_{k} \mathbf{z}_j^{m} \right)
\end{equation}

In this case, $\sigma$ represents a nonlinear activation function. Finally, the outputs of the $K$ attention heads are concatenated to produce the final node representation for each brain region~\cite{velivckovic2017graph}:
\begin{equation}
    \mathbf{h}_i^{m} = \big\|_{k=1}^{K} \mathbf{h}_i^{(k,m)} \in \mathbb{R}^{d_n}
\end{equation}

Assume that $d_n = 64$ refers to the dimensionality of the resulting node embeddings. Accordingly, node embeddings for a specific modality $m$ can be expressed as:
 $\mathbf{H}^{(m)} = \{\mathbf{h}_1^{(m)}, \mathbf{h}_2^{(m)}, \ldots, \mathbf{h}_N^{(m)}\} \in \mathbb{R}^{N \times d_n}$, where $N$ is the number of ROIs. The modality-specific node embedding matrices $\mathbf{H}^{s}$ and $\mathbf{H}^{f}$ represent unified graph-based representations of brain regions that are subsequently utilized for multimodal fusion and downstream classification.

\subsection{Multimodal Fusion Learning}
The multimodal learning framework intends to capture coordinated structural–functional brain changes associated with MDD through the integration of graph-based representations derived from sMRI and rs-fMRI. For systematically investigating the effectiveness of multimodal integration, two fusion strategies are examined: a feature-level concatenation module and a dual cross-attention fusion module.

\subsubsection{Feature-Level Concatenation}
\label{Feature_Concat}
The feature-level concatenation module offers a straightforward and effective baseline for multimodal fusion by directly combining structural and functional graph representations at the embedding level. In this module, the complementary contribution of features derived from sMRI and rs-fMRI are assessed without explicitly modeling cross-modal interactions.

In feature-level fusion, graph-level embeddings are first constructed from the node representations learned by the unified GAT encoder. Assume that $\mathbf{H}^s \in \mathbb{R}^{N \times d_n}, \mathbf{H}^f \in \mathbb{R}^{N \times d_n}$ are the structural and functional node embeddings obtained from the unified GAT encoder, respectively, for a given subject. Specifically, $N$ denotes the number of nodes (ROIs) and $d_n$ is the dimensionality of the node embeddings. Graph-level embeddings are obtained by applying a global average pooling (GAP) operation across the node dimension~\cite{cui2022braingb}: 
\begin{equation}
\mathbf{G}^{s} = \text{GAP}(\mathbf{H}^s) =\frac{1}{N} \sum_{i=1}^{N} \mathbf{H}^{s}_{i}, 
\qquad
\mathbf{G}^{f} = \text{GAP}(\mathbf{H}^f) = \frac{1}{N} \sum_{i=1}^{N} \mathbf{H}^{f}_{i}
\label{eq:GAP}
\end{equation}

Here, $\mathbf{G}^s \in \mathbb{R}^{d_G}, \mathbf{G}^f\in \mathbb{R}^{d_G}$ represent the graph embeddings for the structural and functional modalities,respectively, $d_G = 64$ reflects the dimensionality of the learned structural and functional graph embeddings, and $N$ is the number of nodes (ROIs).

The multimodal representation is then constructed by concatenating the two graph embeddings, as expressed in the following equation:
\begin{equation}\label{eq:Concat}
    \mathbf{Z}_{\text{fusion}} = [\mathbf{G}^s \parallel \mathbf{G}^f] 
\end{equation}

This fused representation preserves modality-specific information while facilitating joint learning during the subsequent classification process. In this formulation, $\parallel$ indicates vector concatenation, while $\mathbf{Z}_\text{fusion} \in \mathbb{R}^{2 \times d_G}$ signifies the fused multimodal graph embeddings.

The multimodal embedding $\mathbf{Z}_\text{fusion}$ is then fed into a multilayer perceptron (MLP) classifier to perform MDD prediction. The MLP consists of two fully connected (FC) layers with nonlinear activation function and dropout regularization. The first FC layer transforms the fused embedding into a hidden feature space in the following manner:
\begin{equation}
    \mathbf{H} = \text{ReLU}\left( \mathbf{W}_1 \mathbf{Z}_{\text{fusion}} + \mathbf{b}_1 \right)
\end{equation}

Assume that the Rectified Linear Unit (ReLU) is a nonlinear function, $\mathbf{W}_1 \in \mathbb{R}^{d_h \times 2 d_G}$ and $\mathbf{b}_1 \in \mathbb{R}^{d_h}$ denote the weight matrix and bias of the first FC layer, and $d_h = 64$ represents the hidden dimension of the MLP.

To improve generalization and reduce overfitting, dropout function is applied:
\begin{equation}
    \tilde{\mathbf{H}} = \text{Dropout}(\mathbf{H})
\end{equation}

Afterward, the second FC layer produces the final prediction labels $\hat{\mathbf{y}}$ as follows:
\begin{equation}
    \hat{\mathbf{y}} = \mathbf{W}_2 \tilde{\mathbf{H}} + \mathbf{b}_2,
\end{equation}

Here, $\mathbf{W}_2 \in \mathbb{R}^{C \times d_h}$, and $\mathbf{b}_2 \in \mathbb{R}^{C}$ are the weight matrix and bias of the second FC layer, and $C$ is the number of classes (MDD vs. HC). The predicted labels $\hat{\mathbf{y}}$ are then passed to a softmax function during training to compute class probabilities $\mathbf{p}$:
\begin{equation}
    \mathbf{p} = \text{softmax}(\hat{\mathbf{y}}).
\end{equation}

A stratified 10-fold cross-validation approach is employed to train and evaluate the proposed model to ensure robust and unbiased performance estimation. Performance metrics are averaged across all folds to provide a reliable assessment of the model's generalization capability. Moreover, we optimized the model using a L2 regularization technique called weight decay in conjunction with the adaptive Moment Estimation (Adam) algorithm. The detailed configuration of this model in our experiments can be found in Table~\ref{tab:Concat}.

\begin{table}[htbp]
\caption{Configuration of the Feature-Level Concatenation Model}  
\centering
\renewcommand{\arraystretch}{1}

\begin{tabular}{|c|c|}

\hline

\textbf{Parameter} & \textbf{Value} \\
\hline
Number of layers & 2 \\
\hline
Hidden dimension & 64 \\
\hline
Output dimension & 2 \\
\hline
Learning rate & $1 \times 10^{-2}$ \\
\hline
Weight decay & $1 \times 10^{-4}$ \\
\hline
Dropout & 0.5 \\
\hline
Batch size & 16 \\
\hline
Optimizer & Adam \\
\hline
Loss function & Cross-entropy \\
\hline
Folds & 10 \\
\hline

\end{tabular}

\label{tab:Concat}
\end{table}

\subsubsection{Dual Cross-Attention Fusion}
To explicitly model interactions between structural and functional brain representations, we present a dual cross-attention fusion module that performs bidirectional attention between node embeddings derived from sMRI and rs-fMRI graphs. Specifically, this module is composed of two complementary components: sMRI-guided rs-fMRI attention, with structural node embeddings attending to functional node embeddings, and rs-fMRI-guided sMRI attention, with functional node embeddings attending to structural node embeddings~\cite{zheng2023attention}. In contract to feature-level concatenation that integrates structural and functional information by directly fusing their graph embeddings, the proposed dual cross-attention approach allows direct information exchange at the node level, permitting each modality to selectively emphasize informative regions in the other modality. Afterward, the updated node embeddings from each modality are aggregated through global pooling to produce graph representations, which are then fused and provided to a MLP for subject-level classification.

Suppose that $\mathbf{H}^s \in \mathbb{R}^{N \times d_n}, \mathbf{H}^f \in \mathbb{R}^{N \times d_n}$ are the node embeddings of the structural and functional graphs, respectively, derived from the unified GAT encoder, where $N$ is the number of ROIs defined by the atlas, and $d_n = 64 $ is the size of the node embeddings for both modalities after projection by the unified GAT encoder.

In the sMRI-guided rs-fMRI attention branch, structural node embeddings guide the refinement of functional representations. In particular, functional node embeddings are projected into queries ($\mathbf{Q}$), while structural node embeddings are projected into keys ($\mathbf{K}$) and values ($\mathbf{V}$) as follows \cite{jimale2025graph}:
\begin{equation}\label{eq:sMRIAtten}
    \mathbf{Q}^f = \mathbf{H}^f \mathbf{W}_Q^f, \quad \mathbf{K}^s = \mathbf{H}^s \mathbf{W}_K^s, \quad \mathbf{V}^s = \mathbf{H}^s \mathbf{W}_V^s
\end{equation}

Where $\mathbf{W}_Q^f, \mathbf{W}_K^s, \mathbf{W}_V^s \in \mathbb{R}^{d_h \times d_a}$ are learnable projection matrices, and $d_a$ is the attention subspace dimension. The attention output is computed using scaled dot-product attention~\cite{jimale2025graph}:
\begin{equation}\label{eq:sMRITOfMRI}
\mathbf{Attention}_{s \rightarrow f} = \mathrm{softmax}\!\left( \frac{\mathbf{Q}^f \left(\mathbf{K}^s\right)^T}{\sqrt{d_a}} \right) \mathbf{V}^s 
\end{equation}

Where $(.)^T$ represents the matrix transpose. The resulting attended features are then integrated with the original functional node embeddings using a residual connection followed by layer normalization (LN)~\cite{jimale2025graph}:
\begin{equation}\label{eq:Res1}
    \mathbf{H'}^f = \mathrm{LN}\!\left( \mathbf{H}^f + \mathrm{Attention}_{s \rightarrow f} \right).
\end{equation}

Similarly, the rs-fMRI-guided sMRI attention branch enables functional node embeddings to guide the refinement of structural representations. In this case, structural node embeddings are projected into queries, whereas functional node embeddings are projected into keys and values~\cite{jimale2025graph}:

\begin{equation}\label{eq:fMRIAtten}
    \mathbf{Q}^s = \mathbf{H}^s \mathbf{W}_Q^s, \quad \mathbf{K}^f = \mathbf{H}^f \mathbf{W}_K^f, \quad \mathbf{V}^f = \mathbf{H}^f \mathbf{W}_V^f
\end{equation}

Assume that $\mathbf{W}_Q^s, \mathbf{W}_K^f, \mathbf{W}_V^f \in \mathbb{R}^{d_h \times d_a}$ refers to the weight matrices of the linear
projections, and $d_a$ reflects the attention subspace dimension. Scaled dot-product attention is then used to calculate the attention output~\cite{jimale2025graph}:
\begin{equation}\label{eq:fMRITOsMRI}
\mathbf{Attention}_{f \rightarrow s} = \mathrm{softmax}\!\left( \frac{\mathbf{Q}^s \left(\mathbf{K}^f \right)^T}{\sqrt{d_a}} \right) \mathbf{V}^f 
\end{equation}

Afterwards, a residual connection is used to combine the attended structural features with their original structural node embeddings, which is subsequently processed by layer normalization (LN)~\cite{jimale2025graph}:
\begin{equation}\label{eq:Res2}
    \mathbf{H'}^s= \mathrm{LN}\!\left( \mathbf{H}^s + \mathrm{Attention}_{f \rightarrow s} \right).
\end{equation}

After bidirectional cross-attention, the refined node embeddings $\mathbf{H'}^s$ and $\mathbf{H'}^f$ are independently aggregated by the GAP function to obtain graph embeddings $\mathbf{G^s}$ and $\mathbf{G^f}$. Finally, the two graph representations are concatenated to create a unified multimodal embedding $\mathbf{Z}_\text{fusion}$, which is then passed to the MLP classifier for subject-level prediction. Notably, the structure of this MLP is the same as that described in Section~\ref{Feature_Concat}.

In this study, a stratified 10-fold cross-validation analysis is conducted to evaluate the proposed dual cross-attention fusion model. Final results are reported as the average of all folds, which reduces partitioning variability and improves reliability. A summary of the detailed training settings and architectural hyperparameters of the dual cross-attention model is presented in Table~\ref{tab:DualAtten}.

\begin{table}[htbp]
\caption{Configuration of the Dual Cross-Attention Fusion Model}  
\centering
\renewcommand{\arraystretch}{1}

\begin{tabular}{|c|c|}

\hline

\textbf{Parameter} & \textbf{Value} \\
\hline
Embedding dimension & 64 \\
\hline
Number of attention heads & 8 \\
\hline
Classification head &  multilayer perceptron (MLP)  \\
\hline
Learning rate & $5 \times 10^{-4}$ \\
\hline
Weight decay & $3 \times 10^{-5}$ \\
\hline
Dropout & 0.3 \\
\hline
Batch size & 16 \\
\hline
Optimizer & Adam \\
\hline
Loss function & Cross-entropy \\
\hline
Folds & 10 \\
\hline

\end{tabular}

\label{tab:DualAtten}
\end{table}

\section{Results and Discussion}
This section outlines the evaluation metrics that we used to evaluate the classification task. Next, we present and discuss the prediction performance of our model using the REST-meta-MDD dataset.
\subsection{Evaluation Metrics}
The entire study is conducted using PyTorch and training is performed on an NVIDIA A100 GPU-equipped Aziz Supercomputer, which significantly speeds up the training of deep learning models for MRI images and graphs, in particular. During evaluation, we assessed model performance using several standard evaluation metrics, including accuracy, sensitivity, specificity, precision, and F1-score. These metrics are calculated as follows:

\begin{equation} 
 \label{eqAcc} 
Accuracy =\frac{(TP + TN)}{(TP + TN + FP + FN)}
\end{equation}
\begin{equation} \label{eqSn}
Sensitivity = \frac{TP}{(TP + FN)} 
\end{equation}
\begin{equation} \label{eqSp}
Specificity = \frac{TN}{(FP + TN)} 
\end{equation}
\begin{equation} \label{eqP}
Precision = \frac{TP}{(TP + FP)}
\end{equation}
\begin{equation} \label{eqF1}
F1-score = \frac{(2 \times Precision \times Sensitivity)}{(Precision + Sensitivity)}
\end{equation}

For this study, $TP$ represents the correct classification of positive samples (MDD), $TN$ represents the correct classification of negative samples (HC), $FP$ reflects the incorrect classification of negative samples as positives, and $FN$ reflects the incorrect classification of positive samples as negatives.

\subsection{Experimental Results}
\subsubsection{Performance Comparison of Fusion Mechanisms}
To measure the effectiveness of different multimodal fusion strategies, we performed a comprehensive comparison between the dual cross-attention mechanism and the widely used feature-level concatenation approach. These two fusion strategies were applied to the same structural and functional graph representations and evaluated using 10-fold stratified cross-validation. This study examined accuracy, sensitivity, specificity, precision, and F1-score to ensure a fair and robust comparison. Table~\ref{tab:Fusion_models_performance} presents the average classification performance and standard deviation for both fusion strategies under different atlas configurations.

For the structural atlases (AAL and HO), feature-level concatenation and dual cross-attention achieved highly comparable performance across all evaluation metrics. For the AAL-based model, feature-level concatenation delivered an accuracy of 75.74$\pm$0.19\%, slightly higher than that obtained using Dual Cross-Attention 75.54$\pm$0.31\%. Both fusion approaches showed similar sensitivity 80.25\%. However, the concatenation method produced marginally higher specificity (70.92$\pm$0.39\% vs. 70.52$\pm$1.21\%), precision (74.62$\pm$0.25\% vs. 74.38$\pm$0.62\%), and F1-score (77.33$\pm$0.14\% vs. 77.20$\pm$0.23\%) compared with dual cross-attention. 
With the HO-based model, feature-level concatenation achieved a slightly higher accuracy of 78.53 $\pm$0.29\%, as compared to dual cross-attention of 78.34$\pm$0.28\%. However, dual cross-attention had a marginally higher sensitivity (75.68$\pm$0.96\% vs. 75.31$\pm$0.00\%), whereas concatenation produced slightly higher specificity (81.97$\pm$0.60\% vs. 81.18$\pm$1.03\%), precision (81.66$\pm$0.50\% vs. 81.10$\pm$0.69\%), and F1-score (78.36$\pm$0.23\% vs. 78.29$\pm$0.36\%). Overall, this high alignment of performance indicates that the dual cross-attention mechanism can effectively integrate structural and functional representations without compromising performance, when using anatomically defined parcellations. This confirms the robustness and stability of the proposed fusion strategy across different atlases. 

In contrast, dual cross-attention consistently outperforms feature-level concatenation across all evaluation metrics for the functional atlases (Dose and CK). In the Dose-based model, dual cross-attention increased accuracy from 81.98$\pm$0.58\% to 83.44$\pm$0.58\%, sensitivity from 80.99$\pm$0.82\% to 83.58$\pm$1.57\%, precision from 83.57$\pm$0.61\% to 84.21$\pm$0.80\%, and F1-score from 82.26$\pm$0.59\% to 83.89$\pm$0.92\%. For the CK-based model, dual cross-attention produced slightly higher accuracy (77.58$\pm$0.79\% vs. 76.82$\pm$1.07\%), sensitivity (77.53$\pm$0.79\% vs. 76.42$\pm$1.61\%), precision (77.63$\pm$1.76\% vs. 77.24$\pm$2.36\%), and F1-score (78.10$\pm$0.81\% vs. 77.28$\pm$0.98\%) compared with concatenation. These consistent improvements in performance prove that the proposed dual cross-attention mechanism is especially effective for functional atlases, in which regions are identified based on functional connectivity patterns. Further, this approach aligns complementary structural and functional information more effectively than direct feature-level concatenation, since it explicitly models cross-modal interactions.  

A further validation of the robustness of the fusion strategies can be found in Table~\ref{tab:Best_MultiFold}. This table presents the classification results for the single fold that had the highest accuracy during cross-validation. Across all atlas configurations, the results demonstrate that feature-level concatenation and dual cross-attention provide comparable performance for structural atlases (AAL and HO), whereas dual cross-attention consistently produces superior performance for functional atlases (Dose and CK). In summary, these findings offer robust evidence that dual cross-attention offers a powerful and expressive technique for multimodal fusion across diverse atlas types when compared with conventional feature-level concatenation.
\begin{table}[htp]
\centering
\caption{Average Classification Results and Standard Deviation of 10-fold Cross-validation for Multimodal Fusion Models. Bolded values indicate the best performing results.}
\label{tab:Fusion_models_performance}
\renewcommand{\arraystretch}{1.8}
\resizebox{\linewidth}{!}{
  \begin{tabular}{|cc|c|c|c|c|c|}\hline 
    \multicolumn{2}{|c}{\multirow{2}{*}{\textbf{Method}}} & 
    \multicolumn{5}{|c|}{ \centering \textbf{Average} $\pm$ \textbf{Standard deviation (\%)} } \\
     \cline{3-7} 
     & & \centering \textbf{Accuracy} & \centering \textbf{Sensitivity} & \centering 
     \textbf{Specificity} & \centering \textbf{Precision} & \centering\arraybackslash 
     \textbf{F1-score} \\
     \hline 
     \multicolumn{1}{|c}{\multirow{4.57}{*}{ \rotatebox[origin=c]{90}{Concatenation}}}& 
     \multicolumn{1}{|c}{\centering AAL-based model} & 
     \multicolumn{1}{|c}{ \begin{tabular}{c}75.74$\pm$0.19 \end{tabular}} & 
     \multicolumn{1}{|c}{\begin{tabular}{c}80.25$\pm$0.00\end{tabular}} & \multicolumn{1}{|c}{\begin{tabular}{c}70.92$\pm$0.39 \end{tabular}} & 
     \multicolumn{1}{|c}{\begin{tabular}{c} 74.62$\pm$0.25\end{tabular}} & 
     \multicolumn{1}{|c|}{\begin{tabular}{c}77.33$\pm$0.14 \end{tabular}}\\ 
     \cline{2-7} 
     & \multicolumn{1}{|c}{\centering HO-based model} & \multicolumn{1}{|c}{\begin{tabular}{c} 78.53$\pm$0.29\end{tabular}} & \multicolumn{1}{|c}{\begin{tabular}{c} 75.31$\pm$0.00\end{tabular}} & \multicolumn{1}{|c}{\begin{tabular}{c}81.97$\pm$0.60\end{tabular}} & \multicolumn{1}{|c}{\begin{tabular}{c}81.66$\pm$0.50\end{tabular}} & \multicolumn{1}{|c|}{\begin{tabular}{c} 78.36$\pm$0.23\end{tabular}}\\
     \cline{2-7} 
     &  \multicolumn{1}{|c}{\centering Dose-based model} & \multicolumn{1}{|c}{\begin{tabular}{c}81.98$\pm$0.58\end{tabular}} & \multicolumn{1}{|c}{\begin{tabular}{c}80.99$\pm$0.82\end{tabular}} & \multicolumn{1}{|c}{\begin{tabular}{c}83.02$\pm$0.71\end{tabular}} & \multicolumn{1}{|c}{\begin{tabular}{c}83.57$\pm$0.61\end{tabular}} & \multicolumn{1}{|c|}{\begin{tabular}{c}82.26$\pm$0.59\end{tabular}}\\ 
     \cline{2-7} 
     & \multicolumn{1}{|c}{\centering CK-based model} & \multicolumn{1}{|c}{\begin{tabular}{c}  76.82$\pm$1.07\end{tabular}} & \multicolumn{1}{|c}{\begin{tabular}{c} 76.42$\pm$1.61\end{tabular}} & \multicolumn{1}{|c}{\begin{tabular}{c}77.24$\pm$2.36\end{tabular}} & \multicolumn{1}{|c}{\begin{tabular}{c}78.20$\pm$1.61\end{tabular}} & \multicolumn{1}{|c|}{\begin{tabular}{c} 77.28$\pm$0.98\end{tabular}}\\ 
     \hline 
     \multicolumn{1}{|c}{\multirow{4.3}{*}{ \rotatebox[origin=c]{90}{Dual cross-attention}}}& 
     \multicolumn{1}{|c}{\centering AAL-based model} &  
     \multicolumn{1}{|c}{\begin{tabular}{c}75.54$\pm$0.31 \end{tabular}} & \multicolumn{1}{|c}{\begin{tabular}{c} 80.25$\pm$0.87\end{tabular}} & 
     \multicolumn{1}{|c}{\begin{tabular}{c}  70.52$\pm$1.21\end{tabular}} & 
     \multicolumn{1}{|c|}{\begin{tabular}{c} 74.38$\pm$0.62\end{tabular}} &
     \multicolumn{1}{|c|}{ \begin{tabular}{c} 77.20$\pm$0.23\end{tabular}} \\
     \cline{2-7} 
     & \multicolumn{1}{|c}{\centering HO-based model} & \multicolumn{1}{|c}{\begin{tabular}{c}  78.34$\pm$0.28\end{tabular}} & \multicolumn{1}{|c}{\begin{tabular}{c} 75.68$\pm$0.96\end{tabular}} & \multicolumn{1}{|c}{\begin{tabular}{c}81.18$\pm$1.03\end{tabular}} & \multicolumn{1}{|c}{\begin{tabular}{c}81.10$\pm$0.69\end{tabular}} & \multicolumn{1}{|c|}{\begin{tabular}{c} 78.29$\pm$0.36\end{tabular}}\\
     \cline{2-7} 
     &  \multicolumn{1}{|c}{\centering Dose-based model} & \multicolumn{1}{|c}{\begin{tabular}{c}\textbf{83.44$\pm$0.58}\end{tabular}} & \multicolumn{1}{|c}{\begin{tabular}{c}\textbf{83.58$\pm$1.57}\end{tabular}} & \multicolumn{1}{|c}{\begin{tabular}{c}\textbf{83.29$\pm$1.03}\end{tabular}} & \multicolumn{1}{|c}{\begin{tabular}{c}\textbf{84.21$\pm$0.80}\end{tabular}} & \multicolumn{1}{|c|}{\begin{tabular}{c}\textbf{83.89$\pm$0.92}\end{tabular}}\\ 
     \cline{2-7} 
     & \multicolumn{1}{|c}{\centering CK-based model} & \multicolumn{1}{|c}{\begin{tabular}{c} 77.58$\pm$0.79\end{tabular}} & \multicolumn{1}{|c}{\begin{tabular}{c} 77.53$\pm$1.54\end{tabular}} & \multicolumn{1}{|c}{\begin{tabular}{c}77.63$\pm$1.76\end{tabular}} & \multicolumn{1}{|c}{\begin{tabular}{c}78.72$\pm$1.18\end{tabular}} & \multicolumn{1}{|c|}{\begin{tabular}{c} 78.10$\pm$0.81\end{tabular}}\\ 
     \hline
  \end{tabular}
}
\end{table}

\begin{table}[htp]
\caption{Classification Results for the Fold with the Highest Accuracy from 10-Fold Stratified Cross-Validation. Bolded values indicate the best performing results \\}
\label{tab:Best_MultiFold}
\resizebox{\columnwidth}{!}{%
\centering
\renewcommand{\arraystretch}{1.8}
  \begin{tabular}{|cc|c|c|c|c|c|}\hline 
    \multicolumn{2}{|c|}{\multirow{1}{*}{\centering \textbf{Method}}} &  
    \centering \textbf{Accuracy\%} & \centering \textbf{Sensitivity\%} & \centering 
     \textbf{Specificity\%} & \centering \textbf{Precision\%} & \centering\arraybackslash 
     \textbf{F1-score\%} \\
     \hline 
     \multicolumn{1}{|c}{\multirow{4.3}{*}{ \rotatebox[origin=c]{90}{Concatenation}}}& 
     \multicolumn{1}{|c}{\centering AAL-based model} & 
     \multicolumn{1}{|c}{ \begin{tabular}{c}75.80 \end{tabular}} & 
     \multicolumn{1}{|c}{\begin{tabular}{c}80.25 \end{tabular}} & \multicolumn{1}{|c}{\begin{tabular}{c} 71.05\end{tabular}} & 
     \multicolumn{1}{|c}{\begin{tabular}{c}74.71 \end{tabular}} & 
     \multicolumn{1}{|c|}{\begin{tabular}{c}77.38\end{tabular}}\\ 
      \cline{2-7} 
     & \multicolumn{1}{|c}{\centering HO-based model} & \multicolumn{1}{|c}{\begin{tabular}{c}78.98  \end{tabular}} & \multicolumn{1}{|c}{\begin{tabular}{c}75.31\end{tabular}} & \multicolumn{1}{|c}{\begin{tabular}{c}82.89\end{tabular}} & \multicolumn{1}{|c}{\begin{tabular}{c}82.43 \end{tabular}} & \multicolumn{1}{|c|}{\begin{tabular}{c}78.71 \end{tabular}}\\
     \cline{2-7} 
     &  \multicolumn{1}{|c}{\centering Dose-based model} & \multicolumn{1}{|c}{\begin{tabular}{c} 82.17\end{tabular}} & \multicolumn{1}{|c}{\begin{tabular}{c}81.48\end{tabular}} & \multicolumn{1}{|c}{\begin{tabular}{c}82.89\end{tabular}} & \multicolumn{1}{|c}{\begin{tabular}{c}83.54\end{tabular}} & \multicolumn{1}{|c|}{\begin{tabular}{c}82.50\end{tabular}}\\ 
     \cline{2-7} 
     & \multicolumn{1}{|c}{\centering CK-based model} & \multicolumn{1}{|c}{\begin{tabular}{c}77.71\end{tabular}} & \multicolumn{1}{|c}{\begin{tabular}{c}76.54\end{tabular}} & \multicolumn{1}{|c}{\begin{tabular}{c}78.95 \end{tabular}} & \multicolumn{1}{|c}{\begin{tabular}{c}79.49\end{tabular}} & \multicolumn{1}{|c|}{\begin{tabular}{c}77.99 \end{tabular}}\\ 
     \hline 
     \multicolumn{1}{|c}{\multirow{4.3}{*}{ \rotatebox[origin=c]{90}{Dual cross-attention}}}& 
     \multicolumn{1}{|c}{\centering AAL-based model} & 
     \multicolumn{1}{|c}{ \begin{tabular}{c}  75.80 \end{tabular}} & 
     \multicolumn{1}{|c}{\begin{tabular}{c}80.25 \end{tabular}} & \multicolumn{1}{|c}{\begin{tabular}{c}71.05  \end{tabular}} & 
     \multicolumn{1}{|c}{\begin{tabular}{c} 74.71\end{tabular}} & 
     \multicolumn{1}{|c|}{\begin{tabular} {c}77.38\end{tabular}}\\ 
      \cline{2-7} 
     & \multicolumn{1}{|c}{\centering HO-based model} & \multicolumn{1}{|c}{\begin{tabular}{c}78.98  \end{tabular}} & \multicolumn{1}{|c}{\begin{tabular}{c}75.31\end{tabular}} & \multicolumn{1}{|c}{\begin{tabular}{c}82.89\end{tabular}} & \multicolumn{1}{|c}{\begin{tabular}{c}82.43\end{tabular}} & \multicolumn{1}{|c|}{\begin{tabular}{c} 78.71\end{tabular}}\\
     \cline{2-7} 
     &  \multicolumn{1}{|c}{\centering Dose-based model} & \multicolumn{1}{|c}{\begin{tabular}{c}\textbf{84.71}\end{tabular}} & \multicolumn{1}{|c}{\begin{tabular}{c}\textbf{86.42}\end{tabular}} & \multicolumn{1}{|c}{\begin{tabular}{c}\textbf{82.89}\end{tabular}} & \multicolumn{1}{|c}{\begin{tabular}{c}\textbf{84.34} \end{tabular}} & \multicolumn{1}{|c|}{\begin{tabular}{c}\textbf{85.37}\end{tabular}}\\ 
     \cline{2-7} 
     & \multicolumn{1}{|c}{\centering CK-based model} & \multicolumn{1}{|c}{\begin{tabular}{c}78.34\end{tabular}} & \multicolumn{1}{|c}{\begin{tabular}{c}79.01\end{tabular}} & \multicolumn{1}{|c}{\begin{tabular}{c}77.63\end{tabular}} & \multicolumn{1}{|c}{\begin{tabular}{c}79.01 \end{tabular}} & \multicolumn{1}{|c|}{\begin{tabular}{c} 79.01 \end{tabular}}\\ 
    \hline
  \end{tabular}%
}
\end{table}

\subsubsection{Comparison with Existing Studies}
Table~\ref{tab:ExistingStudies} presents a comparison of the proposed multimodal framework with representative state-of-the-art approaches for MDD classification based on neuroimaging data, including both single-modality and multimodal approaches. The single-modality methods utilize either sMRI~\cite{alotaibi20263dvit} or rs-fMRI~\cite{alotaibi2025multi}, while multimodal approaches integrate information from both modalities~\cite{zheng2023attention, yuan2023cross, chen2025mmdd, fan2025classifying, li2024evolutionary}. All compared studies were evaluated on the same REST-meta-MDD dataset, allowing fair comparison between different approaches. Notably, we re-implemented both unimodal sMRI~\cite{alotaibi20263dvit} and rs-fMRI~\cite{alotaibi2025multi} models using the same preprocessing pipeline adopted in the proposed framework to ensure consistent experimental settings.

Among all compared methods, the proposed multimodal Dose-based framework provided the best overall performance. It attained the highest accuracy (83.44$\pm$0.58\%), specificity (83.29$\pm$1.03\%), and precision (84.21$\pm$0.80\%). The model also achieved high sensitivity (83.58$\pm$1.57\%) and F1-scores (83.89$\pm$0.92\%), suggesting a balanced classification performance. These results support the robustness and stability of the proposed fusion strategy. Yuan et al.~\cite{yuan2023cross} developed a dynamic cross-domain attention network for fusing multimodal information. They reported the highest sensitivity (87.80$\pm$0.62\%) and F1-score (86.90$\pm$0.53\%) among the compared studies. Nevertheless, overall classification accuracy (81.67$\pm$0.66\%) was lower than that of the proposed method. Furthermore, the relatively low specificity (68.70$\pm$0.47\%) indicates that there is a trade-off between sensitivity and false-positive control. In contrast, Fan et al.~\cite{fan2025classifying} designed a multimodal federated learning framework with the aim of improving data privacy and cross-site generalization. Although they achieved a competitive accuracy of 79.07$\pm$0.02\%, other evaluation metrics like sensitivity and F1-score were not reported, limiting a comprehensive performance comparison. 

Following this, the multimodal HO-based model achieved 78.34$\pm$0.28\% accuracy, with 81.18$\pm$1.03\% specificity, and 81.10$\pm$0.69\% precision, indicating stable and reliable classification performance. In addition, the balanced performance across sensitivity (75.68$\pm$0.96\%) and F1-score (78.29$\pm$0.36\%) further illustrates the robustness of the proposed dual cross-attention mechanism when applied to structurally defined parcellations. In comparison, Chen et al.~\cite{chen2025mmdd} proposed a multimodal multitask disentanglement framework to mitigate site effects and enhance robustness. The model provided an accuracy of 77.76$\pm$1.02\%, with sensitivity of 83.23$\pm$7.06\%, and an F1-score of 79.92$\pm$1.91\%.

Similarly, the multimodal CK-based model yielded 77.58$\pm$0.79\% accuracy, 77.53$\pm$1.54\% sensitivity, 77.63$\pm$1.76\% specificity, 78.72$\pm$1.18\% precision, and 78.10$\pm$0.81\% F1-score. This further demonstrates that the dual cross-attention strategy remains effective across different functional atlas configurations. Li et al.~\cite{li2024evolutionary} developed the M-ENAS framework, an evolutionary neural architecture search method for automated diagnosis of MDD using multimodal MRI imaging. Their method produced an accuracy of 73.68\%, sensitivity of 74.13\%, specificity of 72.88\%, precision of 69.95\%, and F1-score of 71.98\%. Despite producing balanced results, their model generated lower discriminative performance than other multimodal frameworks.

Further comparisons with single-modality models (sMRI-only and rs-fMRI-only) reveal that multimodal integration yielded clear performance improvements for all atlas configurations. As compared to sMRI models, the multimodal framework demonstrated modest improvements for structural atlases (AAL and HO), with accuracy gains of 1.02\% and 0.57\%, along with increases in F1-score of 0.84\% and 0.77\%, respectively. In contrast, the functional atlases were observed to exhibit significantly greater improvements. It was observed that accuracy increased by 7.45\%, specificity by 12.89\%, precision by 9.44\%, and F1-score by 6.16\% for the Dose atlas, while accuracy and F1-score gained by 4.78\% and 4.12\% for the CK atlas.

A more distinct pattern emerged when comparing multimodal models with rs-fMRI models. For structural atlases, accuracy improved by 10.25\% for AAL and 13.95\% for HO, with notable improvements in sensitivity, specificity, precision, and F1-score. There were greater improvements observed for functional atlases, where accuracy increased by 18.47\% for Dose and 12.80\% for CK, alongside significant improvements in sensitivity, specificity, precision, and F1-score.

In summary, the comparative results show that the proposed framework achieves superior and more balanced performance compared to both single-modality and existing multimodal approaches, with substantial gains observed across different atlas configurations and evaluation metrics. These findings support the effectiveness of the proposed multimodal fusion strategy for robust MDD classification by explicitly modeling cross-modal interactions.

\begin{table}[htp]
\centering
\caption{Average Classification Results and Standard Deviation of 10-fold Cross-validation for Existing Multimodal Fusion Models. Bolded values indicate the best performing results.}
\label{tab:ExistingStudies}
\renewcommand{\arraystretch}{1.8}
\resizebox{\linewidth}{!}{
\begin{tabular}{|cc|c|c|c|c|c|}\hline 
    \multicolumn{2}{|c}{\multirow{2}{*}{\textbf{Method}}} & 
    \multicolumn{5}{|c|}{ \centering \textbf{Average} $\pm$ \textbf{Standard deviation (\%)} } \\
     \cline{3-7} 
     & & \centering \textbf{Accuracy} & \centering \textbf{Sensitivity} & \centering 
     \textbf{Specificity} & \centering \textbf{Precision} & \centering\arraybackslash 
     \textbf{F1-score} \\
     \hline 
     \multicolumn{1}{|c}{\multirow{4.57}{*}{ \rotatebox[origin=c]{90}{sMRI models}}}& 
   \multicolumn{1}{|c}{\centering AAL-based model} & 
     \multicolumn{1}{|c}{ \begin{tabular}{c}74.52$\pm$0.90 \end{tabular}} & 
     \multicolumn{1}{|c}{\begin{tabular}{c}79.75$\pm$1.85\end{tabular}} & \multicolumn{1}{|c}{\begin{tabular}{c}68.95$\pm$3.12\end{tabular}} & 
     \multicolumn{1}{|c}{\begin{tabular}{c} 73.31$\pm$1.66\end{tabular}} & 
     \multicolumn{1}{|c|}{\begin{tabular}{c}76.36$\pm$0.64 \end{tabular}}\\ 
     \cline{2-7} 
     & \multicolumn{1}{|c}{\centering HO-based model} & \multicolumn{1}{|c}{\begin{tabular}{c} 77.77$\pm$0.66\end{tabular}} & \multicolumn{1}{|c}{\begin{tabular}{c} 74.32$\pm$2.05\end{tabular}} & \multicolumn{1}{|c}{\begin{tabular}{c}81.45$\pm$1.71\end{tabular}} & \multicolumn{1}{|c}{\begin{tabular}{c}81.06$\pm$1.12\end{tabular}} & \multicolumn{1}{|c|}{\begin{tabular}{c} 77.52$\pm$0.91\end{tabular}}\\
     \cline{2-7} 
     &  \multicolumn{1}{|c}{\centering Dose-based model} & \multicolumn{1}{|c}{\begin{tabular}{c}75.99$\pm$1.18\end{tabular}} & \multicolumn{1}{|c}{\begin{tabular}{c}81.24$\pm$3.70\end{tabular}} & \multicolumn{1}{|c}{\begin{tabular}{c}70.40$\pm$5.80\end{tabular}} & \multicolumn{1}{|c}{\begin{tabular}{c}74.77$\pm$3.08\end{tabular}} & \multicolumn{1}{|c|}{\begin{tabular}{c}77.73$\pm$0.69\end{tabular}}\\ 
     \cline{2-7} 
     & \multicolumn{1}{|c}{\centering CK-based model} & \multicolumn{1}{|c}{\begin{tabular}{c}  72.80$\pm$1.28\end{tabular}} & \multicolumn{1}{|c}{\begin{tabular}{c} 75.06$\pm$3.26\end{tabular}} & \multicolumn{1}{|c}{\begin{tabular}{c}70.40$\pm$3.07\end{tabular}} & \multicolumn{1}{|c}{\begin{tabular}{c}73.05$\pm$1.64\end{tabular}} & \multicolumn{1}{|c|}{\begin{tabular}{c} 73.98$\pm$1.53\end{tabular}}\\ 
     \hline 
     \multicolumn{1}{|c}{\multirow{4.57}{*}{ \rotatebox[origin=c]{90}{rs-fMRI models}}}& 
     \multicolumn{1}{|c}{\centering AAL-based model} & 
     \multicolumn{1}{|c}{ \begin{tabular}{c}65.29$\pm$2.32 \end{tabular}} & 
     \multicolumn{1}{|c}{\begin{tabular}{c}76.67$\pm$7.46\end{tabular}} & \multicolumn{1}{|c}{\begin{tabular}{c}53.16$\pm$10.96\end{tabular}} & 
     \multicolumn{1}{|c}{\begin{tabular}{c} 64.02$\pm$3.50\end{tabular}} & 
     \multicolumn{1}{|c|}{\begin{tabular}{c}69.41$\pm$1.94 \end{tabular}}\\ 
     \cline{2-7} 
     & \multicolumn{1}{|c}{\centering HO-based model} & \multicolumn{1}{|c}{\begin{tabular}{c} 64.39$\pm$1.08\end{tabular}} & \multicolumn{1}{|c}{\begin{tabular}{c} 74.32$\pm$6.91\end{tabular}} & \multicolumn{1}{|c}{\begin{tabular}{c}53.82$\pm$6.48\end{tabular}} & \multicolumn{1}{|c}{\begin{tabular}{c}63.29$\pm$1.43\end{tabular}} & \multicolumn{1}{|c|}{\begin{tabular}{c} 68.15$\pm$2.52\end{tabular}}\\
     \cline{2-7} 
     &  \multicolumn{1}{|c}{\centering Dose-based model} & \multicolumn{1}{|c}{\begin{tabular}{c}64.97$\pm$2.85\end{tabular}} & \multicolumn{1}{|c}{\begin{tabular}{c}70.62$\pm$12.97\end{tabular}} & \multicolumn{1}{|c}{\begin{tabular}{c}58.95$\pm$11.33\end{tabular}} & \multicolumn{1}{|c}{\begin{tabular}{c}65.19$\pm$2.97\end{tabular}} & \multicolumn{1}{|c|}{\begin{tabular}{c}66.93$\pm$6.23\end{tabular}}\\ 
     \cline{2-7} 
     & \multicolumn{1}{|c}{\centering CK-based model} & \multicolumn{1}{|c}{\begin{tabular}{c}  64.78$\pm$2.13\end{tabular}} & \multicolumn{1}{|c}{\begin{tabular}{c} 71.85$\pm$6.62\end{tabular}} & \multicolumn{1}{|c}{\begin{tabular}{c}57.24$\pm$10.25\end{tabular}} & \multicolumn{1}{|c}{\begin{tabular}{c}64.64$\pm$3.53\end{tabular}} & \multicolumn{1}{|c|}{\begin{tabular}{c} 67.72$\pm$1.73\end{tabular}}\\ 
     \hline 
     \multicolumn{1}{|c}{\multirow{7.5}{*}{ \rotatebox[origin=c]{90}{Multimodal models}}}& 
     \multicolumn{1}{|c}{\centering AAL-based model} &  
     \multicolumn{1}{|c}{\begin{tabular}{c}75.54$\pm$0.31 \end{tabular}} & \multicolumn{1}{|c}{\begin{tabular}{c} 80.25$\pm$0.87\end{tabular}} & 
     \multicolumn{1}{|c}{\begin{tabular}{c}  70.52$\pm$1.21\end{tabular}} & 
     \multicolumn{1}{|c|}{\begin{tabular}{c} 74.38$\pm$0.62\end{tabular}} &
     \multicolumn{1}{|c|}{ \begin{tabular}{c} 77.20$\pm$0.23\end{tabular}} \\
     \cline{2-7} 
     & \multicolumn{1}{|c}{\centering HO-based model} & \multicolumn{1}{|c}{\begin{tabular}{c}  78.34$\pm$0.28\end{tabular}} & \multicolumn{1}{|c}{\begin{tabular}{c} 75.68$\pm$0.96\end{tabular}} & \multicolumn{1}{|c}{\begin{tabular}{c}81.18$\pm$1.03\end{tabular}} & \multicolumn{1}{|c}{\begin{tabular}{c}81.10$\pm$0.69\end{tabular}} & \multicolumn{1}{|c|}{\begin{tabular}{c} 78.29$\pm$0.36\end{tabular}}\\
     \cline{2-7} 
     &  \multicolumn{1}{|c}{\centering Dose-based model} & \multicolumn{1}{|c}{\begin{tabular}{c}\textbf{83.44$\pm$0.58}\end{tabular}} & \multicolumn{1}{|c}{\begin{tabular}{c}83.58$\pm$1.57\end{tabular}} & \multicolumn{1}{|c}{\begin{tabular}{c}\textbf{83.29$\pm$1.03}\end{tabular}} & \multicolumn{1}{|c}{\begin{tabular}{c}\textbf{84.21$\pm$0.80}\end{tabular}} & \multicolumn{1}{|c|}{\begin{tabular}{c}83.89$\pm$0.92\end{tabular}}\\ 
     \cline{2-7} 
     & \multicolumn{1}{|c}{\centering CK-based model} & \multicolumn{1}{|c}{\begin{tabular}{c} 77.58$\pm$0.79\end{tabular}} & \multicolumn{1}{|c}{\begin{tabular}{c} 77.53$\pm$1.54\end{tabular}} & \multicolumn{1}{|c}{\begin{tabular}{c}77.63$\pm$1.76\end{tabular}} & \multicolumn{1}{|c}{\begin{tabular}{c}78.72$\pm$1.18\end{tabular}} & \multicolumn{1}{|c|}{\begin{tabular}{c} 78.10$\pm$0.81\end{tabular}}\\ 
     \cline{2-7} 
     & \multicolumn{1}{|c}{\centering Zheng et al. \cite{zheng2023attention}} & \multicolumn{1}{|c}{\begin{tabular}{c} 72.00$\pm$1.70\end{tabular}} & \multicolumn{1}{|c}{\begin{tabular}{c} 64.9$\pm$9.80\end{tabular}} & \multicolumn{1}{|c}{\begin{tabular}{c}77.5$\pm$5.80\end{tabular}} & \multicolumn{1}{|c}{\begin{tabular}{c}\textemdash\end{tabular}} & \multicolumn{1}{|c|}{\begin{tabular}{c} \textemdash\end{tabular}}\\ 
     \cline{2-7} 
     & \multicolumn{1}{|c}{\centering Yuan et al.\cite{yuan2023cross}} & \multicolumn{1}{|c}{\begin{tabular}{c} 81.67$\pm$0.66\end{tabular}} & \multicolumn{1}{|c}{\begin{tabular}{c} \textbf{87.80$\pm$0.62}\end{tabular}} & \multicolumn{1}{|c}{\begin{tabular}{c}68.70$\pm$0.47\end{tabular}} & \multicolumn{1}{|c}{\begin{tabular}{c}\textemdash\end{tabular}} & \multicolumn{1}{|c|}{\begin{tabular}{c} \textbf{86.90$\pm$0.53}\end{tabular}}\\ 
     \cline{2-7} 
     & \multicolumn{1}{|c}{\centering Chen et al.\cite{chen2025mmdd}} & \multicolumn{1}{|c}{\begin{tabular}{c} 77.76$\pm$1.02\end{tabular}} & \multicolumn{1}{|c}{\begin{tabular}{c} 83.23$\pm$7.06\end{tabular}} & \multicolumn{1}{|c}{\begin{tabular}{c}\textemdash\end{tabular}} & \multicolumn{1}{|c}{\begin{tabular}{c}\textemdash\end{tabular}} & \multicolumn{1}{|c|}{\begin{tabular}{c} 79.92$\pm$1.91\end{tabular}}\\ 
     \cline{2-7} 
     & \multicolumn{1}{|c}{\centering Fan et al. \cite{fan2025classifying}} & \multicolumn{1}{|c}{\begin{tabular}{c} 79.07$\pm$0.02\end{tabular}} & \multicolumn{1}{|c}{\begin{tabular}{c} \textemdash\end{tabular}} & \multicolumn{1}{|c}{\begin{tabular}{c}\textemdash\end{tabular}} & \multicolumn{1}{|c}{\begin{tabular}{c}\textemdash\end{tabular}} & \multicolumn{1}{|c|}{\begin{tabular}{c} \textemdash\end{tabular}}\\
     \cline{2-7} 
     & \multicolumn{1}{|c}{\centering Li et al. \cite{li2024evolutionary}} & \multicolumn{1}{|c}{\begin{tabular}{c} 73.68\end{tabular}} & \multicolumn{1}{|c}{\begin{tabular}{c} 74.13\end{tabular}} & \multicolumn{1}{|c}{\begin{tabular}{c}72.88\end{tabular}} & \multicolumn{1}{|c}{\begin{tabular}{c}69.95\end{tabular}} & \multicolumn{1}{|c|}{\begin{tabular}{c} 71.98\end{tabular}}\\
     \hline
  \end{tabular}
}
\end{table}

\subsection{Statistical Comparison Between Single-Modal and Multimodal Approaches }
In this section, two-sample t-tests were conducted to determine whether there were significant differences in performance between multimodal and single-modal models (sMRI-only and rs-fMRI–only). The results are summarized in Table~\ref{tab:Stat_p}. In this study, differences were considered statistically significant when $p < 0.05$

The comparison between the multimodal and sMRI models showed statistically significant improvements in some evaluation metrics across the atlases. The multimodal approach exhibited highly significant gains in accuracy, specificity, precision, and F1-score for the Dose and CK functional atlases, indicating the importance of integrating structural and functional information. For the AAL and HO structural atlases, significant improvements were observed in accuracy and F1-score, while other metrics showed no statistically significant differences.

Similar trends were observed when comparing multimodal with rs-fMRI models. Multimodal approaches produced statistically significant improvements for functional atlases across all evaluated metrics, while improvements for structural atlases were found in accuracy, specificity, precision, and F1-score. 

Overall, the statistical analysis reveals that the proposed multimodal fusion technique consistently outperforms single-modal approaches in multiple atlas configurations. The most significant improvements were observed for functional atlases, while significant gains were also achieved for structural atlases, demonstrating the robustness and effectiveness of explicitly modeling cross-modal interactions.

\begin{table}[htbp]
\caption{Statistical Analysis of Significant Differences (p-values) Between Single-Modal and Multimodal Approaches. Bolded values indicate statistically significant differences between models at $p < 0.05$. \\}
\label{tab:Stat_p}
\resizebox{\columnwidth}{!}{%
\renewcommand{\arraystretch}{1.8}
  \begin{tabular}{|cc|c|c|c|c|c|}\hline 
    \multicolumn{2}{|c|}{\multirow{1}{*}{\centering \textbf{Method}}} &  
    \centering \textbf{Accuracy} & \centering \textbf{Sensitivity} & \centering 
     \textbf{Specificity} & \centering \textbf{Precision} & \centering\arraybackslash 
     \textbf{F1-score} \\
     \hline 
     \multicolumn{1}{|c}{\multirow{4.3}{*}{ Multimodal and sMRI models}}& 
     \multicolumn{1}{|c}{\centering AAL-based model} & 
     \multicolumn{1}{|c}{ \begin{tabular}{c} \textbf{0.0048} \end{tabular}} & 
     \multicolumn{1}{|c}{\begin{tabular}{c} 0.4695 \end{tabular}} & \multicolumn{1}{|c}{\begin{tabular}{c}0.1751\end{tabular}} & 
     \multicolumn{1}{|c}{\begin{tabular}{c} 0.0865 \end{tabular}} & 
     \multicolumn{1}{|c|}{\begin{tabular}{c} \textbf{0.0017}\end{tabular}}\\ 
      \cline{2-7} 
     & \multicolumn{1}{|c}{\centering HO-based model} & \multicolumn{1}{|c}{\begin{tabular}{c} \textbf{0.0294} \end{tabular}} & \multicolumn{1}{|c}{\begin{tabular}{c}0.0890\end{tabular}} & \multicolumn{1}{|c}{\begin{tabular}{c}0.6959\end{tabular}} & \multicolumn{1}{|c}{\begin{tabular}{c}0.9300\end{tabular}} & \multicolumn{1}{|c|}{\begin{tabular}{c}\textbf{0.0295} \end{tabular}}\\
     \cline{2-7} 
     &  \multicolumn{1}{|c}{\centering Dose-based model} & \multicolumn{1}{|c}{\begin{tabular}{c} \textbf{0.0000}\end{tabular}} & \multicolumn{1}{|c}{\begin{tabular}{c}0.0965 \end{tabular}} & \multicolumn{1}{|c}{\begin{tabular}{c}\textbf{0.0000}\end{tabular}} & \multicolumn{1}{|c}{\begin{tabular}{c}\textbf{0.0000}\end{tabular}} & \multicolumn{1}{|c|}{\begin{tabular}{c}\textbf{0.0000}\end{tabular}}\\ 
     \cline{2-7} 
     & \multicolumn{1}{|c}{\centering CK-based model} & \multicolumn{1}{|c}{\begin{tabular}{c}\textbf{0.0000}\end{tabular}} & \multicolumn{1}{|c}{\begin{tabular}{c}0.0547\end{tabular}} & \multicolumn{1}{|c}{\begin{tabular}{c}\textbf{0.0000} \end{tabular}} & \multicolumn{1}{|c}{\begin{tabular}{c}\textbf{0.0000}\end{tabular}} & \multicolumn{1}{|c|}{\begin{tabular}{c}\textbf{0.0000} \end{tabular}}\\ 
     \hline 
     \multicolumn{1}{|c}{\multirow{4.3}{*}{ Multimodal and rs-fMRI models}}& 
     \multicolumn{1}{|c}{\centering AAL-based model} & 
     \multicolumn{1}{|c}{ \begin{tabular}{c} \textbf{0.0000}  \end{tabular}} & 
     \multicolumn{1}{|c}{\begin{tabular}{c} 0.1691 \end{tabular}} & \multicolumn{1}{|c}{\begin{tabular}{c} \textbf{0.0002} \end{tabular}} & 
     \multicolumn{1}{|c}{\begin{tabular}{c} \textbf{0.0000}\end{tabular}} & 
     \multicolumn{1}{|c|}{\begin{tabular} {c}\textbf{0.0000}\end{tabular}}\\ 
      \cline{2-7} 
     & \multicolumn{1}{|c}{\centering HO-based model} & \multicolumn{1}{|c}{\begin{tabular}{c} \textbf{0.0000} \end{tabular}} & \multicolumn{1}{|c}{\begin{tabular}{c}0.5674\end{tabular}} & \multicolumn{1}{|c}{\begin{tabular}{c}\textbf{0.0000} \end{tabular}} & \multicolumn{1}{|c}{\begin{tabular}{c} \textbf{0.0000}\end{tabular}} & \multicolumn{1}{|c|}{\begin{tabular}{c}\textbf{0.0000} \end{tabular}}\\
     \cline{2-7} 
     &  \multicolumn{1}{|c}{\centering Dose-based model} & \multicolumn{1}{|c}{\begin{tabular}{c}\textbf{0.0000}\end{tabular}} & \multicolumn{1}{|c}{\begin{tabular}{c}\textbf{0.0081}\end{tabular}} & \multicolumn{1}{|c}{\begin{tabular}{c}\textbf{0.0000}\end{tabular}} & \multicolumn{1}{|c}{\begin{tabular}{c}\textbf{0.0000} \end{tabular}} & \multicolumn{1}{|c|}{\begin{tabular}{c}\textbf{0.0000}\end{tabular}}\\ 
     \cline{2-7} 
     & \multicolumn{1}{|c}{\centering CK-based model} & \multicolumn{1}{|c}{\begin{tabular}{c}\textbf{0.0000}\end{tabular}} & \multicolumn{1}{|c}{\begin{tabular}{c}\textbf{0.0221}\end{tabular}} & \multicolumn{1}{|c}{\begin{tabular}{c}\textbf{0.0000}\end{tabular}} & \multicolumn{1}{|c}{\begin{tabular}{c}\textbf{0.0000} \end{tabular}} & \multicolumn{1}{|c|}{\begin{tabular}{c} \textbf{0.0000} \end{tabular}}\\ 
    \hline
  \end{tabular}%
}

\end{table}

\section{Conclusion}
In this study, we proposed a multimodal graph-based framework for MDD classification that explicitly models structural–functional interactions through a dual cross-attention mechanism. The proposed approach offers reciprocal refinement of modality-specific embeddings while preserving graph topology by performing bidirectional cross-attentions between corresponding nodes of the structural and functional brain graphs. In this manner, complementary multimodal information can be utilized more effectively, thereby optimizing discriminative representation learning at the node level. A comprehensive analysis of the large-scale REST-meta-MDD dataset has demonstrated that the proposed framework achieves robust and balanced performance across multiple atlas configurations. The model consistently enhances classification metrics compared to both single-modality and existing multimodal approaches while maintaining stability under the stratified cross-validation method. These findings emphasize the importance of explicitly modeling bidirectional structural–functional dependencies for multimodal neuroimaging analysis.

Several research directions remain to be explored in the future. First, extending the framework toward multi-atlas integration may further enhance representational diversity by capturing complementary anatomical and functional parcellation schemes within a unified learning framework. Second, incorporating additional imaging modalities, such as DTI may improve the richness of multimodal representations. Finally, integrating explainability mechanisms would provide significant neurobiological insights into modality-specific and region-specific contributions.

\bibliographystyle{ieeetr}
\setcitestyle{square}
\bibliography{main}

\end{document}